\documentclass[10pt,twocolumn,letterpaper]{article}

\usepackage{cvpr}              

\usepackage[dvipsnames]{xcolor}

\definecolor{cvprblue}{rgb}{0.21,0.49,0.74}
\usepackage[pagebackref,breaklinks,colorlinks,citecolor=cvprblue]{hyperref}
\usepackage{multirow}
\usepackage{threeparttable}
\usepackage{xcolor}
\usepackage{colortbl}
\usepackage{color}
\usepackage{makecell}
\usepackage{float}

\def\confName{CVPR}
\def\confYear{2024}

\title{Weakly Supervised Monocular 3D Detection with a Single-View Image}

\author{
Xueying Jiang$^{1}$ \quad
Sheng Jin$^1$ \quad
Lewei Lu$^2$ \quad
Xiaoqin Zhang$^3$ \quad 
Shijian Lu$^1$\thanks{Corresponding author.} \quad
\\
    $^1$S-Lab, Nanyang Technological University \quad $^2$Sensetime Research \\
    $^3$College of Computer Science and Technology, Zhejiang University of Technology \\ \\
}

\begin{document}
\maketitle
\begin{abstract}
Monocular 3D detection (M3D) aims for precise 3D object localization from a single-view image which usually involves labor-intensive annotation of 3D detection boxes. Weakly supervised M3D has recently been studied to obviate the 3D annotation process by leveraging many existing 2D annotations, but it often requires extra training data such as LiDAR point clouds or multi-view images which greatly degrades its applicability and usability in various applications. We propose SKD-WM3D, a weakly supervised monocular 3D detection framework that exploits depth information to achieve M3D with a single-view image exclusively without any 3D annotations or other training data. One key design in SKD-WM3D is a self-knowledge distillation framework, which transforms image features into 3D-like representations by fusing depth information and effectively mitigates the inherent depth ambiguity in monocular scenarios with little computational overhead in inference. In addition, we design an uncertainty-aware distillation loss and a gradient-targeted transfer modulation strategy which facilitate knowledge acquisition and knowledge transfer, respectively. Extensive experiments show that SKD-WM3D surpasses the state-of-the-art clearly and is even on par with many fully supervised methods.
\end{abstract}    
\section{Introduction}

Monocular 3D detection (M3D) has emerged as one key component in the area of autonomous driving and computer vision. Its primary target is to recognize objects and obtain their 3D localization from single-view images. Thanks to its low deployment cost, M3D~\cite{chen2016monocular, peng2022weakm3d} has attracted increasing attention in both academic and industrial sectors, achieving very impressive progress in recent years. On the other hand, most existing studies~\cite{ku2019monocular, simonelli2020disentangling, reading2021categorical, peng2022did} adopt a fully supervised setup which have been facing increasing scalability concern as large-scale 3D boxes are often labor-intensive to collect. Effective M3D training without 3D annotations has become a critical issue while handling M3D problems in various research and practical tasks.

\begin{figure*}[htbp]
	\centering
	\includegraphics[width=\linewidth]{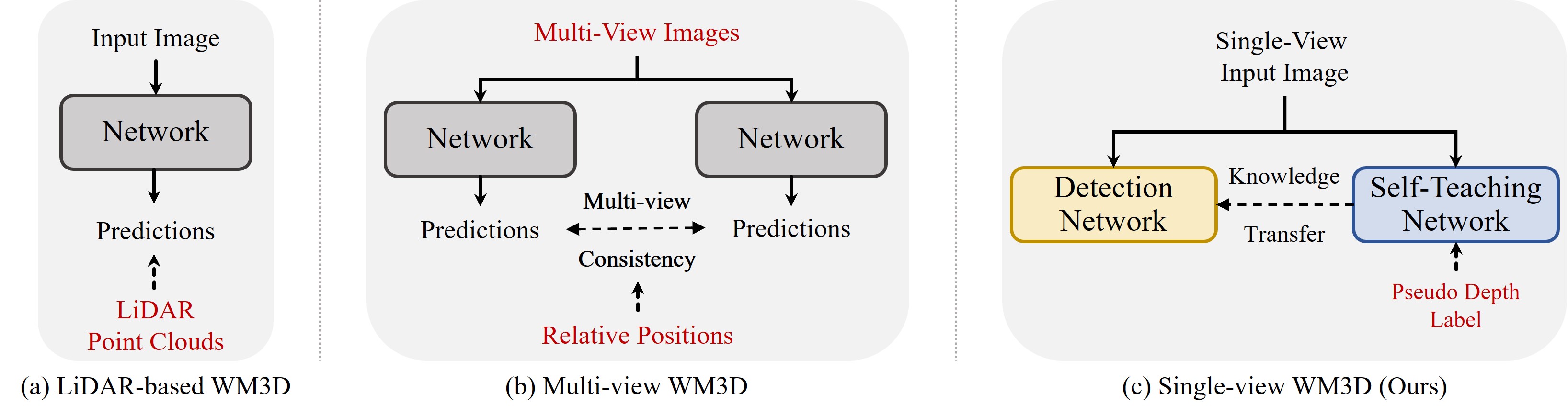}
    
	\caption{Different paradigms in weakly supervised monocular 3D detection. Our approach in (c) leverages \textit{Pseudo Depth Labels} from a single-view image to achieve weakly supervised monocular 3D detection, requiring no extra training data like LiDAR point clouds or multi-view images as in (a) and (b). It improves usability and applicability greatly. 
    The \textit{Pseudo Depth Labels} are obtained with an off-the-shelf depth estimator~\cite{hu2021penet} without extra training and ground-truth depth labels.
    Data in \textcolor[RGB]{192,0,0}{red} denotes extra data in network training.
    }
	\label{fig:introduction}
\end{figure*}

Weakly supervised M3D (WM3D)~\cite{peng2022weakm3d} has recently been explored for learning effective 3D detectors without 3D box annotations, aiming to exploit 2D annotations to make up for the absence of 3D information. For example, WeakM3D~\cite{peng2022weakm3d} exploits LiDAR point clouds to infer 3D information as illustrated in Figure~\ref{fig:introduction}(a). However, it requires costly and complicated LiDAR sensors to collect point clouds which limits its applicability and usability greatly. WeakMono3D~\cite{tao2023weakly} employs 2D information only by either leveraging multi-view stereo with images from multiple cameras or constructing pseudo-multi-view perspective from sequential video frames as illustrated in Figure~\ref{fig:introduction}(b). However, collecting multi-view images is complicated, and resorting to a pseudo multi-view perspective degrades the detection performance clearly. With the advance of single-view depth estimation, WM3D with depth from a single-view image presents a potential solution for compensating the absence of 3D annotations. On the other hand, direct integration of such depth into existing frameworks often necessitates complex network architectures which further incurs significant computational costs. This gives rise to a pertinent question: When not using additional LiDAR point clouds or multi-view image pairs, is it possible to harness the depth from off-the-shelf depth estimators without introducing much computational overhead in inference?

We design SKD-WM3D, a novel weakly supervised monocular 3D object detection method that is exclusively grounded on single-view images. One key design in SKD-WM3D is a self-knowledge distillation framework which consists of a \textbf{D}epth-guided \textbf{S}elf-teaching \textbf{N}etwork (DSN) and a \textbf{M}onocular 3D \textbf{D}etection \textbf{N}etwork (MDN). As illustrated in Figure~\ref{fig:introduction}(c), SKD-WM3D utilizes depth information obtained from an off-the-shelf depth estimator~\cite{hu2021penet} to enhance the 3D localization ability of DSN and transfers such ability to MDN via self-knowledge distillation. Such self-distillation design enables MDN to unearth the intrinsic depth information from single-view images independently, bypassing additional modules such as pre-trained depth estimation networks and leading to precise and efficient 3D localization with little computational overhead during inference. 
On top of DSN and MDN, we design an uncertainty-aware distillation loss to optimize the utilization of the transferred 3D localization knowledge by weighting up more certain knowledge while weighting down less certain knowledge. In addition, we design a gradient-targeted transfer modulation strategy to synchronize the learning paces of DSN and MDN during the process of learning 3D localization knowledge, by prioritizing MDN learning at the initial stage when MDN lags behind DSN and enabling it to provide more feedback to DSN when MDN is better trained at late stages.

Our contribution can be summarized in three aspects. \textit{First}, we design a novel framework that achieves weakly supervised monocular 3D detection by distilling knowledge between a depth-guided self-teaching network and a monocular 3D detection network. Without any extra training data like LiDAR point clouds or multi-view images, the framework exploits depth exclusively from a single image with little computational overhead in inference. \textit{Second}, we design an uncertainty-aware distillation loss and a gradient-targeted transfer modulation strategy which facilitate knowledge acquisition and knowledge transfer, respectively. \textit{Third}, the proposed approach clearly outperforms the state-of-the-art in weakly supervised monocular 3D detection, and its performance is even on par with several fully supervised methods.

\section{Related Work}


\subsection{Monocular 3D Detection}

Monocular 3D object detection aims to predict 3D object localization from single-view images. Standard monocular detectors~\cite{he2019mono3d, brazil2019m3d, chen2020monopair, zhou2021monoef, zhang2023monodetr} operate solely on single images, without utilizing additional data. However, the inherent depth ambiguity of monocular detection significantly hinders its performance compared to its stereo counterparts. To address this limitation, various approaches seek solutions with the help of extra data, such as LiDAR point clouds~\cite{ku2019monocular, ma2019accurate, chen2021monorun, chong2022monodistill}, video sequences~\cite{brazil2020kinematic}, 3D CAD models~\cite{chen2016monocular, liu2021autoshape, murthy2017reconstructing}, and depth estimation~\cite{ding2020learning, qin2019monogrnet, wang2019pseudo, you2019pseudo}. Specifically, MonoRUn~\cite{chen2021monorun} adopts an uncertainty-aware regional reconstruction network for regressing pixel-associated 3D object coordinates with LiDAR point clouds as extra supervision. MonoDistill~\cite{chong2022monodistill} introduces an effective distillation-based approach that incorporates spatial information from LiDAR signals into monocular 3D detection. Additionally, pseudo-LiDAR-based methods~\cite{wang2019pseudo, you2019pseudo} convert estimated depth maps to simulate the real LiDAR point clouds to utilize the well-designed LiDAR-based 3D detector. During inference, compared with methods using depth estimation, our method eliminates the need for pseudo depth labels and complex network architectures, with little computational overhead. Besides, existing fully supervised methods require large-scale 3D box ground truth, which is labor-intensive to collect and annotate.
\begin{figure*}[t]
	\centering
	\includegraphics[width=1\linewidth]{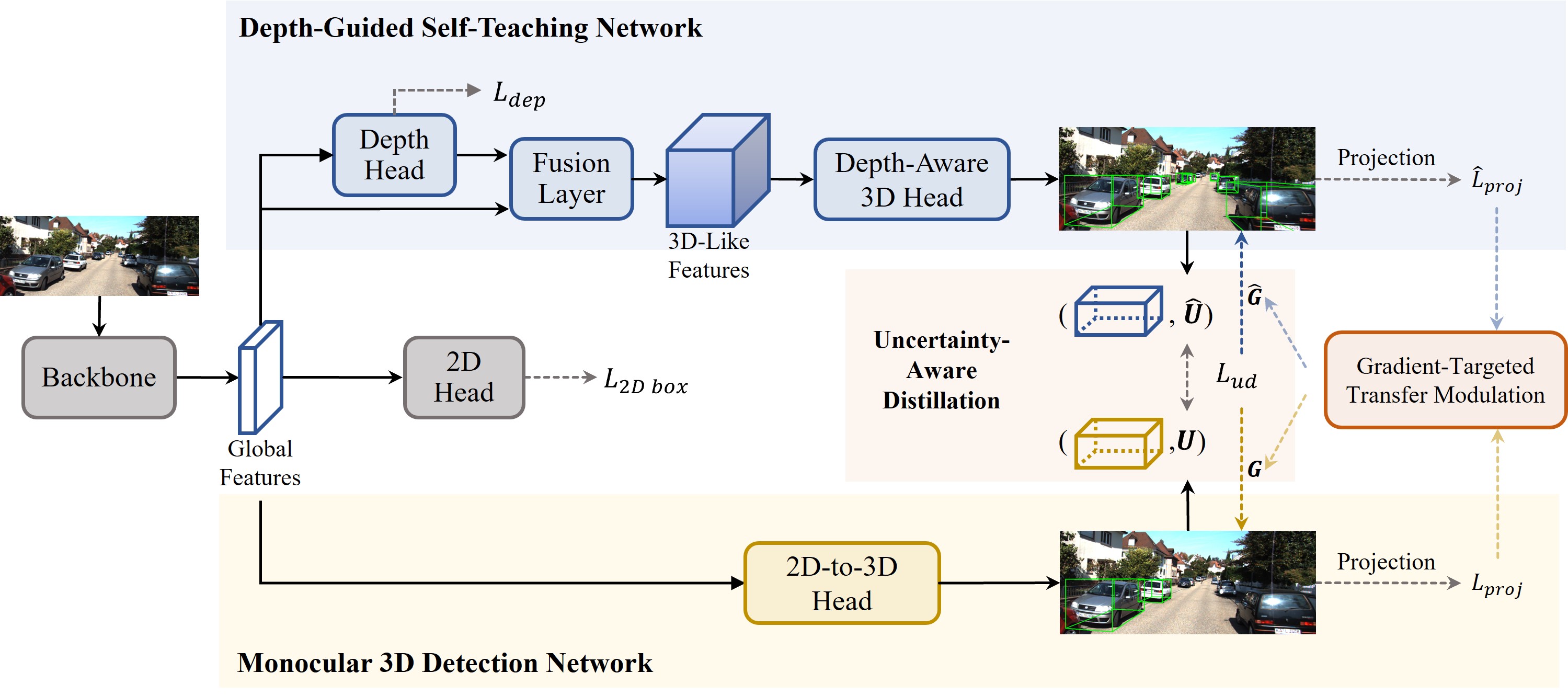}

	\caption{The framework of the proposed self-knowledge distillation network. The framework consists of a depth-guided self-teaching network and a monocular 3D detection network. The depth-guided self-teaching network acquires comprehensive 3D localization knowledge by leveraging depth information and transfers its learned expertise to the monocular 3D detection network via soft label distillation to enhance its performance. We design an uncertainty-aware distillation loss and a gradient-targeted transfer modulation strategy to facilitate the knowledge transfer between the two networks effectively. During inference, the monocular 3D detection network extracts intrinsic depth information from single-view images independently with little computational overhead.
 }
	\label{fig:overall_architecture}
\end{figure*}

\subsection{Weakly Supervised 3D Object Detection}
Due to the high cost of annotating 3D boxes in the 3D object detection task, various weakly supervised approaches have been proposed.
For example, WS3D~\cite{meng2020weakly} presents a weakly supervised method for 3D LiDAR object detection, which requires only a limited number of weakly annotated scenes with center-annotated BEV maps. VS3D~\cite{qin2020weakly} introduces a cross-model knowledge distillation strategy to transfer the knowledge from the RGB domain to the point cloud domain, using LiDAR point clouds as weak supervision. Recent research on weakly supervised 3D object detection has turned to explore the monocular setting. For example, WeakM3D~\cite{peng2022weakm3d} generates 2D boxes to select RoI LiDAR point clouds as weak supervision and then predicts 3D boxes that closely align with the selected RoI LiDAR point clouds. 
More recently, WeakMono3D~\cite{tao2023weakly} eliminates the need for LiDAR, offering both multi-view and single-view yet multi-frame versions. While the former acquires stereo image inputs from multiple cameras, the latter constructs a pseudo-multi-view perspective using multiple video frames. The multi-frame version exhibits inferior 3D scene comprehension compared with the multi-view approach due to its smaller inter-frame disparity, leading to degraded performance.
Instead of requiring extra training data like LiDAR point clouds or multi-view images, we tackle the challenge of weakly supervised monocular 3D detection by leveraging a single-view image exclusively.


\subsection{Self-Knowledge Distillation}
Knowledge distillation~\cite{hinton2015distilling, liu2019knowledge, park2019relational, tian2019contrastive, romero2014fitnets, chung2020feature, zhao2022decoupled, huang2022knowledge, gao2023dqs3d} aims to transfer knowledge from a pre-trained teacher network to a student network for improving its performance.
Self-knowledge distillation~\cite{szegedy2016rethinking, muller2019does, yang2023knowledge}, distinct from traditional knowledge distillation, leverages the information within the student network to facilitate its learning without the pre-trained teacher network. Specifically, data augmentation approach~\cite{xu2019data, yun2020regularizing, heo2019comprehensive} transfers knowledge through different distortions of the same training data. However, they are susceptible to inappropriate augmentations, such as improper instance rotation or distortion, potentially introducing noise that hampers network learning. Another typical approach exploits auxiliary networks~\cite{zhu2018knowledge, zhang2019your}. For example, DKS~\cite{sun2019deeply} introduces auxiliary supervision branches and pairwise knowledge alignments, while FRSKD~\cite{ji2021refine} adds a new branch supervised by the original features and utilizes both soft-label and feature-map distillation. Our work is the first that introduces self-knowledge distillation with auxiliary networks for weakly supervised monocular 3D detection. It effectively exploits depth information from single-view images with little computational overhead during inference.

\section{Methodology}

This section presents the proposed SKD-WM3D. First, the problem definition and overview are presented in Sec.~\ref{subsec:problem_definition}. Then detailed designs of SKD-WM3D are introduced, including the self-knowledge distillation framework in Sec.~\ref{subsec:skd_framework_architecture}, the uncertainty-aware distillation loss in Sec.~\ref{subsec:uncertainty_based_consistency_loss} and the gradient-targeted transfer modulation strategy in Sec.~\ref{subsec:transfer_modulate_strategy}. Finally, loss functions are presented in Sec.~\ref{subsec:loss_function}. 

\subsection{Problem Definition and Overview \label{subsec:problem_definition}}
Weakly supervised monocular 3D detection takes an RGB image and the corresponding 2D bounding boxes as supervision, aiming to classify objects and determine their bounding boxes in 3D space without involving any 3D annotations in training. The prediction of each object is composed of the object category $C$, a 2D bounding box $B^{2D}$, and a 3D bounding box $B^{3D}$. Specifically, the 3D box $B^{3D}$ can be further decomposed to the object 3D location $(x_{3D}, y_{3D}, z_{3D})$, the object dimension with height, width and length $(h_{3D}, w_{3D}, l_{3D})$, as well as orientation $\theta$.


We design a self-knowledge distillation framework to tackle the challenge of weakly supervised monocular 3D detection from a single-view image. As Figure~\ref{fig:overall_architecture} shows, the framework consists of two subnetworks including a \textit{Depth-Guided Self-Teaching Network} and a \textit{Monocular 3D Detection Network}. In the \textit{Depth-Guided Self-Teaching Network}, the global features $F_G$ extracted by the backbone are fed into a \textit{Depth Head} to obtain depth features. Next, the global features $F_G$ and the extracted depth features are fed into a \textit{Fusion Layer} to obtain 3D-like features $F_{3D}$. The 3D-like features of each object are then obtained via RoIAlign, and further fed to a \textit{Depth-Aware 3D Head} to predict 3D box $\widehat{B}^{3D}_{p}$ and uncertainty $\widehat{U}$. In the \textit{Monocular 3D Detection Network}, We first use RoIAlign to generate object-level features from the global features $F_G$, and then feed them to a \textit{2D-to-3D Head} to predict 3D box $\widehat{B}^{2D}_{p}$ and uncertainty $U$. Besides, the 3D boxes predicted by both networks are further projected into 2D boxes. Moreover, we design an uncertainty-aware distillation loss $L_{ud}$ to obtain low-uncertainty knowledge, and a gradient-targeted transfer modulation strategy to synchronize the learning paces between the two networks by controlling gradients $\widehat{G}$ and $G$ of $L_{ud}$.


\subsection{Self-Knowledge Distillation Framework \label{subsec:skd_framework_architecture}}

The self-knowledge distillation framework enhances the 3D localization ability of the depth-guided self-teaching network by utilizing depth information from an off-the-shelf depth estimator and then transfers the ability to the monocular 3D detection network via self-knowledge distillation.

\paragraph{Depth-Guided Self-Teaching Network.} To equip the self-teaching network with 3D localization ability, we propose to learn from global features $F_G$ and depth information from an off-the-shelf depth estimator to acquire comprehensive 3D knowledge. The depth information is exploited via two major designs. Firstly, we introduce a depth head $\mathcal{D}$ that extracts depth features $F_D$ as follows:

\begin{equation}
    F_D=\mathcal{D}(F_G),
\end{equation}

The depth features $F_D$ are exploited to generate depth maps $D_p$, where the depth map generation is supervised by the pseudo ground truth of the depth map $D_{gt}$ that is predicted by an off-the-shelf depth estimator by using the focal loss~\cite{lin2017focal} as depth loss $L_{dep}$. Hence, the depth features can be acquired by the depth-guided self-teaching network effectively.

\paragraph{Remark 1.} We generate depth pseudo labels using an off-the-shelf depth estimator~\cite{hu2021penet} with frozen weights, eliminating the need for additional training and ground-truth depth labels. Adopting an off-the-shelf depth estimator incurs negligible costs as compared with prior studies that require either LiDAR point clouds~\cite{peng2022weakm3d} or multi-view images~\cite{tao2023weakly}. 



Secondly, we obtain 3D-like features $F_{G3D}$ by integrating the depth features $F_D$ that provide information along the depth dimension, as well as the global features $F_G$ that capture knowledge about the 2D image plane. Specifically, we design a fusion layer that fuses the depth features $F_D$ with the global features $F_G$ to derive the $F_{G3D}$ as follows:

\begin{equation}
    F_{G3D}=FFN(CA(SA(F_D), F_G)),
\end{equation}
where the $FFN$ is the feed-forward network, and $CA$, $SA$ denote $CrossAttention$, $SelfAttention$, respectively. The structures of $CrossAttention$ and $SelfAttention$ employ the standard transformer architecture~\cite{vaswani2017attention}. The obtained 3D comprehension improves the network's ability to precisely locate objects, effectively mitigating depth ambiguity arising from single-view image input.

\paragraph{Monocular 3D Detection Network.} The monocular 3D detection network acquires the 3D localization knowledge from the depth-guided self-teaching network. By distilling soft labels generated by the depth-guided self-teaching network, the monocular 3D detection network can extract intrinsic depth information from images independently during inference. This kills the need for additional complex modules such as pre-trained depth estimation networks or depth fusion modules, facilitating the inference with little computational overhead.




\subsection{Uncertainty-Aware Distillation Loss \label{subsec:uncertainty_based_consistency_loss}}
During the knowledge distillation process, uncertain knowledge could affect the network training negatively if all transferred knowledge is treated equally. To benefit more from certain knowledge and weaken the effect of uncertain knowledge, we design an uncertainty-aware distillation loss between the 3D boxes that are predicted by the two networks in the self-knowledge distillation framework. The uncertainty-aware distillation loss exploits the prediction uncertainty to modulate the distillation loss magnitude as follows:


\begin{equation}
    L_{ud}=\frac{L_{d}}{min((\widehat{U}+U)/2, \alpha)}+\left \| min(\frac{\widehat{U}+U}{2}, \alpha)  \right \|^{2},  
    \label{equation_2}
\end{equation}

where $\widehat{U}$ and $U$ denote the uncertainty of the 3D boxes that are predicted by the two networks, respectively. Here we assume the 3D box predictions have the Laplace distribution, and we adopt the standard deviations as the uncertainties, inspired by~\cite{lu2021geometry,kendall2017uncertainties, peng2022did}. $\left \| min(\frac{\widehat{U}+U}{2}, \alpha)  \right \|^{2}$ is the L2 regularization, and $\alpha$ is a fixed value set to 0.1. $L_{d}$ denotes the basic distillations loss, and we employ the commonly used SmoothL1~\cite{girshick2015fast} loss to enforce the consistency between the 3D boxes predicted by the two networks. The SmoothL1 loss leaves a soft margin when computing the difference between the two 3D boxes:

\begin{small}
\begin{equation}
L_{d}=\left\{\begin{array}{ll}
0.5 \times (\widehat{B}^{3D}_{p}-B^{3D}_{p})^2, & \text { if }|\widehat{B}^{3D}_{p}-B^{3D}_{p}|<1.0 \\
|\widehat{B}^{3D}_{p}-B^{3D}_{p}|-0.5, & \text { otherwise }
\end{array}\right.,
\label{equation_1}
\end{equation}
\end{small}

where $\widehat{B}^{3D}_{p}$ and $B^{3D}_{p}$ are the predicted 3D boxes from the depth-guided self-teaching network and the monocular 3D detection network, respectively. 

\paragraph{Remark 2.} The proposed uncertainty-aware distillation loss $L_{ud}$ integrates average uncertainty $\frac{\widehat{U}+U}{2}$ as regularization and a weighted component for the basic distillation loss $L_{d}$, allowing adaptive learning adjustments based on the knowledge's uncertainty level. Specifically, when dealing with uncertain knowledge, a smaller weight is assigned to the basic distillation loss $L_{d}$ to mitigate potential adverse effects on network learning. Consequently, the network prioritizes optimizing uncertainty reduction in such scenarios. When dealing with certain knowledge, the network emphasizes optimizing the basic distillation loss $L_{d}$ due to its higher weight. Notably, the basic distillation loss $L_{d}$ simply considers box consistency, while integrating uncertainty is beneficial for enhancing the knowledge distillation process.


\subsection{Transfer Modulation Strategy \label{subsec:transfer_modulate_strategy}}
The depth-guided self-teaching network, which leverages depth information to predict 3D boxes, transfers its learned 3D knowledge to the monocular 3D detection network. The asynchronous learning paces of the two networks pose potential challenges to effective 3D knowledge transfer.



We design a gradient-targeted transfer modulation strategy to synchronize the learning pace of the depth-guided self-teaching network and the monocular 3D detection network. We modulate the knowledge transfer dynamically, by controlling the gradients from the uncertainty-aware distillation loss $L_{ud}$. Specifically, we adapt the gradients based on the 2D projection performance of each network, assigning smaller backward gradients for the good-performing network and higher backward gradients for the bad-performing network. The gradient-targeted transfer modulation strategy is formulated as follows:
\begin{equation}
    \widehat{G}' = \frac{2 \times \widehat{L}_{proj}}{\widehat{L}_{proj} + L_{proj}} \times \widehat{G}, 
    G' = \frac{2 \times L_{proj}}{\widehat{L}_{proj} + L_{proj}} \times G
    \label{equation_3},
\end{equation}

Where $\widehat{G}$ and $G$ are the original gradients of the two networks, $\widehat{G}'$ and $G'$ are the modified gradients, $\widehat{L}_{proj}$ and $L_{proj}$ are projection losses, computed between the projected 2D boxes from 3D box predictions and 2D box annotations. 

The gradient-targeted transfer modulation strategy prioritizes training the monocular 3D detection network when its learning lags behind the depth-guided self-teaching network at the early training stage. As the monocular 3D detection network learns and improves gradually, it is enabled to provide more feedback progressively to the depth-guided self-teaching network.


\subsection{Loss Functions \label{subsec:loss_function}} 

The overall objective consists of three losses including $L_{ud}$, $L_{dep}$ and $L_{base}$. $L_{ud}$ is the uncertainty-aware distillation loss as defined in Sec.~\ref{subsec:uncertainty_based_consistency_loss}. $L_{dep}$ is the depth loss for supervising the predicted depth map. $L_{base}$ includes losses for supervising 2D boxes prediction by 2D heads and the 3D box predictions, which has been adopted in prior CenterNet~\cite{zhou2019objects} and WeakMono3D~\cite{tao2023weakly}. We set the weight for each loss item to 1.0, and the overall loss function can be formulated as follows:
\begin{equation}
    L = L_{ud} + L_{dep} + L_{base}
    \label{equation_6}.
\end{equation}

\section{Experiments}
\subsection{Datasets}
We conduct experiments over the KITTI 3D dataset~\cite{geiger2012we} and the nuScenes dataset~\cite{caesar2020nuscenes} that have been widely adopted for benchmarking of 3D object detection methods. The KITTI 3D dataset consists of 7,481 images for training and 7,518 images for testing. The labels of the train set are publicly available and the labels of the test set are stored on a test server for evaluation. For ablation studies, we follow~\cite{chen2016monocular} which divides the 7,481 training samples into a new train set with 3,712 images and a validation set with 3,769 images. The nuScenes dataset comprises 1,000 video scenes, including RGB images captured by 6 surround-view cameras. The dataset is split into a training set (700 scenes), a validation set (150 scenes), and a test set (150 scenes). Following~\cite{peng2022weakm3d}, the performance on the validation set is reported.

\begin{table*}[ht]
\centering
\renewcommand\arraystretch{1.05}
\setlength\tabcolsep{12pt}
\scalebox{1.1}{
\begin{tabular}{l|c|c|ccc}
\hline
\multirow{2}{*}{Method} & \multirow{2}{*}{Backbone} & \multirow{2}{*}{Supervision}  & \multicolumn{3}{c}{AP$_{BEV}/$AP$_{3D}   ($IoU$=0.7)|_{R_{40}}$} \\
\cline{4-6} 
                        &        &                                           & Easy                  & Moderate            & Hard               \\ \hline\hline
WeakM3D~\cite{peng2022weakm3d}      & ResNet-50           & \multirow{3}{*}{Weak}                        & 11.82/5.03            & 5.66/2.26           & 4.08/1.63          \\
WeakMono3D~\cite{tao2023weakly}   & DLA-34           &                                          &  12.31/6.98            &  8.80/4.85           &  7.81/4.45          \\

SKD-WM3D (Ours)                   &    DLA-34 &                                      &  \cellcolor[HTML]{EFEFEF}    \textbf{15.71/8.95}               & \cellcolor[HTML]{EFEFEF}   \textbf{10.15/5.54}             & \cellcolor[HTML]{EFEFEF} \textbf{8.08/4.53 }                 \\ \hline
\end{tabular}
}
\caption{
    Comparison on the performance of the Car category on KITTI \textit{test} set. For all results, we use AP$|_{R_{40}}$ metrics with IoU threshold equal to 0.7. The best results are in \textbf{bold}.
}
\label{tab:table_1_kitti_test_car_0.7}
\end{table*}


\begin{table*}[ht]

\centering
\renewcommand\arraystretch{1.05}
\setlength\tabcolsep{12pt}
\scalebox{1.01}{
\begin{threeparttable}
\begin{tabular}{l|c|c|ccc}
\hline
\multirow{2}{*}{Method} & \multirow{2}{*}{Backbone} & \multirow{2}{*}{Supervision}  & \multicolumn{3}{c}{AP$_{BEV}/$AP$_{3D}   ($IoU$=0.5)|_{R_{40}}$} \\ 
\cline{4-6} 
                        &  &                                                    & Easy                 & Moderate            & Hard                \\ \hline\hline
CenterNet~\cite{zhou2019objects}   & DLA-34            & \multirow{9}{*}{Full}         & 34.36/20.00          & 27.91/17.50         & 24.65/15.57         \\
MonoGRNet~\cite{qin2019monogrnet}    & VGG-16           &                                               & 52.13/47.59          &  35.99/32.28         &  28.72/25.50         \\
M3D-RPN~\cite{brazil2019m3d}   & DenseNet-121              &                                               & 53.35/48.53          & 39.60/35.94         & 31.76/28.59         \\
MonoPair~\cite{chen2020monopair}    & DLA-34            &                                                &  61.06/55.38          &  47.63/42.39         &  41.92/37.99         \\
MonoDLE~\cite{ma2021delving}     & DLA-34            &                                              & 60.73/55.41          & 46.87/43.42         & 41.89/37.81         \\
GUPNet~\cite{lu2021geometry}    & DLA-34              &                                              & 61.78/57.62          &  47.06/42.33         &  40.88/37.59         \\
Kinematic~\cite{brazil2020kinematic} & DenseNet-121                   &                            &        61.79/55.44     
   & 44.68/39.47       &  34.56/31.26     \\
MonoDistill~\cite{chong2022monodistill}   & DLA-34                  &                                        & 71.45/65.69           &  53.11/49.35          & 46.94/43.49         \\
MonoDETR~\cite{zhang2023monodetr}*   & ResNet-50            &                                              & 72.34/68.05           & 51.97/48.42        &  46.94/43.48          \\
\hline

VS3D~\cite{qin2020weakly}   & VGG-16                 & \multirow{5}{*}{Weak}                     & 31.59/22.62          & 20.59/14.43         & 16.28/10.91         \\
Autolabels~\cite{zakharov2020autolabeling} & ResNeXt101             &                                              &  50.51/38.31          &  30.97/19.90         &  23.72/14.83         \\
WeakM3D~\cite{peng2022weakm3d}    & ResNet-50             &                                               & \textbf{58.20}/50.16          & 38.02/29.94         & 30.17/23.11         \\
WeakMono3D~\cite{tao2023weakly}  & DLA-34            &                                                &  54.32/49.37          &  42.83/39.01         &  40.07/36.34         \\

SKD-WM3D (Ours)     & DLA-34              &                                     &   
\cellcolor[HTML]{EFEFEF}
55.47/\textbf{50.21}                   & \cellcolor[HTML]{EFEFEF} \textbf{44.35/41.57}                   &  \cellcolor[HTML]{EFEFEF} \textbf{41.86/36.92}                   \\ \hline

\end{tabular}

\end{threeparttable}
}
\caption{
    Comparison on the performance of the Car category on KITTI \textit{val} set. For all results, we use AP$|_{R_{40}}$ metric with IoU threshold equal to 0.5. * denotes this performance is reproduced from the official code.
    The best results of weakly supervised 3D object detection approaches are in \textbf{bold}.
}
\label{tab:table_2_kitti_val_car_0.5}
\end{table*}

\begin{table}[ht]
\centering
\renewcommand\arraystretch{1.05}
\scalebox{0.98}{
\begin{tabular}{l|cccc}
\hline
Method    & AP$\uparrow$ & ATE$\downarrow$ &  ASE$\downarrow$  & AAE$\downarrow$                      \\ \hline\hline
WeakM3D~\cite{peng2022weakm3d} & 0.214   & 0.814 & 0.234  & 0.682    \\  
SKD-WM3D (Ours) & \textbf{0.242}  & \textbf{0.795} & \textbf{0.231} & \textbf{0.659}  \\  \hline
\end{tabular}
}
\caption{
    Comparison on the performance of the Car category on nuScenes \textit{val} set. The best results are in \textbf{bold}.
    }
\label{tab:nuscenes}
\end{table}

\subsection{Evaluation Protocols}

For the KITTI 3D dataset, following~\cite{simonelli2020disentangling}, we adopt the evaluation metric AP$|_{R_{40}}$ which is the average of the AP of 40 recall points. We report the average precision on bird’s eye view and 3D object detection as AP$_{BEV}|_{R_{40}}$ and AP$_{3D}|_{R_{40}}$. In addition, as most weakly supervised 3D object detection methods apply IoU threshold of 0.7 for the test set and 0.5 for the validation set, we adopt the same thresholds for fair benchmarking. We adopt four metrics for the evaluation on the nuScenes dataset, namely, AP (average precision), ATE (average translation error), ASE (average scale error), and AAE (average attribute error). Following~\cite{peng2022weakm3d}, AVE (Average Velocity Error) and AOE (Average Orientation Error) are not reported due to the lack of supervision for velocity and movement direction in the weakly supervised approach.

\subsection{Implementation Details}
We conduct experiments on 2 NVIDIA V100 GPUs with batch size of 16, and train the framework with 150 epochs. We use the Adam optimizer with the initial learning rate $1e^{-5}$, which is gradually increased to $1e^{-3}$ for the first 5 epochs and decayed with rate 0.1 at the 90 and 120 epochs. We employ DLA-34~\cite{yu2018deep} as the detector’s backbone. The pseudo ground truth of the depth map is generated with an off-the-shelf depth estimator~\cite{hu2021penet} without using the ground truth of depth label.

\begin{figure*}[t]
    \centering
	\includegraphics[width=1\linewidth]{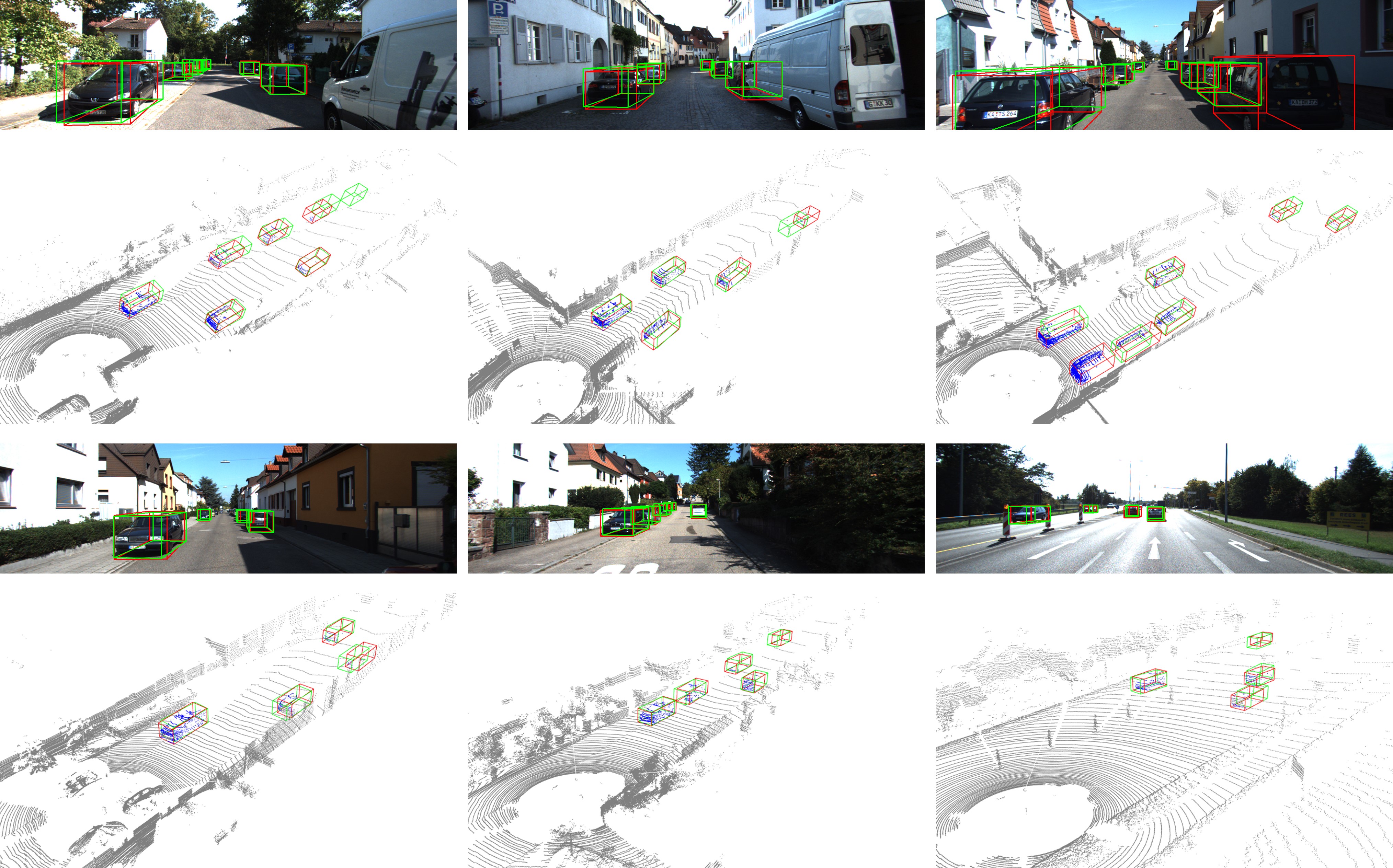}

    \caption{Qualitative illustration on KITTI \textit{val} set. \textcolor{red}{Red} boxes denote ground-truth annotations and \textcolor{green}{Green} boxes denote our predictions. The ground truth of LiDAR point clouds is utilized for visualization purposes only. Best viewed with zoom-in.}
    \label{fig:visualization}
\end{figure*}

\begin{figure*}[t]
    \centering
	\includegraphics[width=1\linewidth]{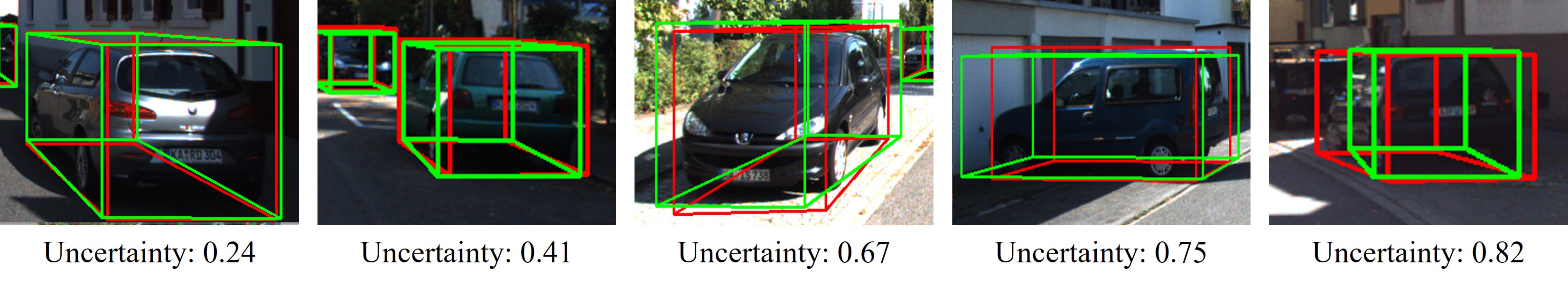}

    \caption{Qualitative illustration of object detection and the corresponding detection uncertainties on KITTI \textit{val} set. \textcolor{red}{Red} boxes denote ground-truth annotations and \textcolor{green}{Green} boxes denote our predictions. The detection accuracy is closely correlated with the detection uncertainty. Best viewed with zoom-in.
    }
    \label{fig:uncertainty}
\end{figure*}

\subsection{Comparison with State-of-the-Art Methods}
We compare our method with several state-of-the-art weakly supervised monocular 3D detection methods on the KITTI test set. As Table~\ref{tab:table_1_kitti_test_car_0.7} shows, our method achieves superior detection performance across all metrics. This superior performance is largely attributed to our designed self-knowledge distillation framework that extracts and exploits intrinsic depth information from a single-view image effectively. It should be highlighted that our method employs a single-view image exclusively without involving additional training data such as LiDAR point clouds~\cite{peng2022weakm3d} or multi-view image pairs~\cite{tao2023weakly}.


Table~\ref{tab:table_2_kitti_val_car_0.5} shows the benchmarking on the KITTI validation set. Specifically, we compare our method against both state-of-the-art weakly supervised monocular 3D detection methods and fully supervised methods. It can be seen that our method achieves superior performance compared to WeakM3D~\cite{peng2022weakm3d} and WeakMono3D~\cite{tao2023weakly} across most metrics, even without using LiDAR point clouds or multi-view images. Notably, our method significantly outperforms WeakM3D~\cite{peng2022weakm3d} in the Moderate and Hard categories, primarily due to the sparse nature of distant LiDAR point clouds adversely affecting its performance. Additionally, its performance is even on par with several fully supervised methods~\cite{zhou2019objects, qin2019monogrnet, brazil2019m3d}.

Table~\ref{tab:nuscenes} shows the results on the nuScenes validation set. It can be seen that our proposed method outperforms WeakM3D~\cite{peng2022weakm3d} across all four evaluation metrics, validating the effectiveness of our approach.



\paragraph{Qualitative Results} 
Figure~\ref{fig:visualization} shows qualitative results with both 2D RGB images and 3D point clouds. In simple scenarios, our model achieves great prediction accuracy, which is largely attributed to the proposed self-knowledge distillation framework as well as the uncertainty-aware distillation loss and the gradient-targeted transfer modulation strategy, all working together to facilitate comprehensive 3D information extraction effectively. However, for heavily occluded or distant objects, the accuracy of orientation and depth estimation tends to drop, which is common for monocular 3D detection due to its ill-posed nature. In addition, Figure~\ref{fig:uncertainty} shows the visualization of predicted bounding boxes and their corresponding uncertainties. It can be observed that the prediction accuracy of bounding boxes has a close correlation with the prediction uncertainty.

\begin{table}[t]
\renewcommand\arraystretch{1.05}
\centering
\scalebox{0.83}{
\begin{tabular}{c|cc|ccc}
\hline
\multirow{2}{*}{Index} & \multirow{2}{*}{\makecell[c]{MDN}} & \multirow{2}{*}{\makecell[c]{DSN}} & \multicolumn{3}{c}{AP$_{BEV}/$AP$_{3D}   ($IoU$=0.5)|_{R_{40}}$} \\ \cline{4-6} 
 &  &  & Easy & Moderate & Hard \\ \hline\hline
1     &    \checkmark                           &            & 0.00/0.00 & 0.00/0.00   &   0.00/0.00      \\ 
2     &         &  \checkmark                      & 45.23/40.96 &  34.27/31.02  & 30.17/26.27 \\
3     & \checkmark        & \checkmark            &   
\textbf{55.47/50.21}                   & \textbf{44.35/41.57}                   &  \textbf{41.86/36.92}        \\ 
\hline
\end{tabular}
}
\caption{
    Ablation study of the proposed self-knowledge distillation framework. The best results are in \textbf{bold}. MDN denotes the Monocular 3D Detection Network, while DSN denotes the Depth-Guided Self-Teaching Network.
    }
\label{tab:ablation_study_branch}
\end{table}

\begin{table}[t]
\renewcommand\arraystretch{1.05}
\centering
\scalebox{0.83}{
\begin{tabular}{c|cc|ccc}
\hline
\multirow{2}{*}{Index} & \multirow{2}{*}{\makecell[c]{$L_{ud}$}} & \multirow{2}{*}{\makecell[c]{TMS} } & \multicolumn{3}{c}{AP$_{BEV}/$AP$_{3D}   ($IoU$=0.5)|_{R_{40}}$} \\ \cline{4-6} 
                       &                            &                            & Easy        & Moderate        & Hard        \\ \hline\hline
                       
1     &                               &     &       49.95/44.61 & 38.24/35.74   &   37.28/34.82         \\ 
2     &    \checkmark                           &     &       53.16/48.13 & 41.85/39.02   &   40.14/35.70         \\ 
3     &         &  \checkmark                      &  52.35/46.30   &  41.45/38.91   &  39.73/35.44    \\
4     & \checkmark        & \checkmark            &   
\textbf{55.47/50.21}                   &  \textbf{44.35/41.57}                   &   \textbf{41.86/36.92}        \\ 
\hline
\end{tabular}
}
\caption{
    Ablation study of the proposed uncertainty-aware distillation loss and the gradient-targeted transfer modulation strategy. The best results are in \textbf{bold}. $L_{ud}$ denotes the Uncertainty-Aware Distillation Loss, while TMS denotes the gradient-targeted transfer modulation strategy.
    }
\label{tab:ablation_study_loss_strategy}
\end{table}

\begin{table}[t]
\renewcommand\arraystretch{1.05}
\centering
\setlength\tabcolsep{3pt}
\scalebox{0.85}{
\begin{tabular}{l|c|c|c}
\hline
Method    & WeakM3D~\cite{peng2022weakm3d}
& MonoDistill~\cite{chong2022monodistill}*                      & SKD-WM3D (Ours)* \\ \hline\hline
FPS & 13.9 & 25.0 & 30.3    \\  \hline
\end{tabular}
}
\caption{
    Comparison on inference speed of M3D methods. * denotes this method utilizes dense depth maps.
    }
\label{tab:ablation_study_inference_time}
\end{table}

\subsection{Ablation Study}

We conduct extensive ablation studies on the KITTI validation dataset to evaluate our designs. Specifically, we evaluated the efficacy of the two individual networks in the proposed self-knowledge distillation framework. Additionally, we examine the effect of the proposed uncertainty-aware distillation loss and the gradient-targeted transfer modulation strategy. Lastly, we evaluated the efficiency of our monocular 3D detection framework.


\paragraph{Self-Knowledge Distillation Framework.} We train two models to assess the contributions of the two networks in our proposed self-knowledge distillation framework. As Table~\ref{tab:ablation_study_branch} shows, training the monocular 3D detection network alone produces few meaningful detection results as the absence of depth information leads to ambiguous object localization along the depth dimension. As a comparison, training the depth-guided self-teaching network alone can produce reasonable detection results thanks to the depth map pseudo labels. In addition, training both subnetworks concurrently produces the best 3D detection, validating the effectiveness of extracting 3D information from a single image. We can also see that including the self-knowledge distillation on top of the depth-guided self-teaching network greatly improves the detection by reducing the adverse effects of uncertain knowledge and enabling communication between the two subnetworks during training.

\paragraph{Uncertainty-Aware Distillation Loss and Gradient-Targeted Transfer Modulation Strategy.} Table~\ref{tab:ablation_study_loss_strategy} shows the ablation study of the proposed uncertainty-aware distillation loss and the gradient-targeted transfer modulation strategy. It can be observed that the baseline does not perform well due to the adverse effect of uncertain knowledge and the asynchronous learning paces of the two subnetworks. On top of the baseline, including either the uncertainty-aware distillation loss or the gradient-targeted transfer modulation strategy improves the detection performance significantly, underscoring the importance of attaining high-certainty knowledge and synchronizing the learning paces of the two networks. In addition, combining the two designs achieves the best performance, highlighting their complementary nature and collaborative roles in knowledge acquisition and knowledge transfer.


\paragraph{Inference speed comparison.}  Table~\ref{tab:ablation_study_inference_time} compares the inference speed on the KITTI validation set. Our method demonstrates superior efficiency thanks to our designed self-knowledge distillation framework, without utilizing complex network architectures during inference.


\section{Conclusion}
In this paper, we point out that previous weakly supervised monocular 3D detection methods either require additional LiDAR point clouds or paired images from multiple viewpoints or temporal sequences. To overcome these constraints, we propose a weakly supervised monocular 3D object detection approach that is exclusively grounded on single-view image inputs. Central to our approach is a self-knowledge distillation framework, which effectively harnesses the depth information within a single-view image with little computational overhead during inference. We further introduce an uncertainty-aware distillation loss and a gradient-targeted transfer modulation strategy, facilitating knowledge acquisition and knowledge transfer, respectively. Finally, extensive experiments demonstrate the effectiveness of our method. Moving forward, we plan to further generalize our work to diverse and challenging scenarios, such as occlusions, varying lighting, and weather conditions, thereby enhancing its practical applicability.

\section*{Acknowledgement}
This study is supported under the RIE2020 Industry Alignment Fund – Industry Collaboration Projects (IAF-ICP) Funding Initiative, as well as cash and in-kind contribution from the industry partner(s).

{
    \small
    \bibliographystyle{ieeenat_fullname}
    \bibliography{main}
}


\end{document}